\documentclass[letterpaper,10pt,journal,twoside]{IEEEtran}
\usepackage{amsmath,amsfonts}
\usepackage{algorithmic}
\usepackage{algorithm}
\usepackage{array}
\usepackage[caption=false,font=normalsize,labelfont=sf,textfont=sf]{subfig}
\usepackage{textcomp}
\usepackage{stfloats}
\usepackage{url}
\usepackage{verbatim}
\usepackage{graphicx}
\usepackage{cite}
\usepackage{booktabs}
\usepackage{amssymb}
\usepackage{multirow}
\usepackage{hyperref}

\begin{document}
\bstctlcite{IEEEexample:BSTcontrol}
\title{Perception-Aware Autonomous Exploration \\ in Feature-Limited Environments}

\author{Moji Shi$^{1}$, Rajitha de Silva$^{2}$, Hang Yu$^{1}$, Riccardo Polvara$^{2}$, and Marija Popovi\'{c}$^{1}$
\thanks{Manuscript received: February 26, 2026; Revised: May 18, 2026; Accepted: June 22, 2026.}
\thanks{This paper was recommended for publication by Editor Ayoung Kim upon evaluation of the Associate Editor and Reviewers’ comments.}
\thanks{$^{1}$M. Shi, H. Yu, and M. Popovi\'{c} are with the MAVLab, TU Delft, Netherlands. $^{2}$R. de Silva and R. Polvara are with the Lincoln Center for Autonomous Systems (L-CAS), School of Engineering and Physical Sciences, University of Lincoln, UK.}
\thanks{Digital Object Identifier (DOI): see top of this page.}}

\markboth{IEEE Robotics and Automation Letters. Preprint Version. Accepted June, 2026}
{Shi \MakeLowercase{\textit{et al.}}: Perception-Aware Autonomous Exploration}


\maketitle

\begin{abstract}
  Autonomous exploration in unknown environments relies on onboard state estimation for localisation and mapping. Existing exploration methods primarily maximise coverage efficiency, but often overlook that visual-inertial odometry (VIO) performance strongly depends on the availability of robust visual features. Consequently, exploration policies can drive a robot into feature-sparse regions where tracking degrades, leading to odometry drift, corrupted maps, and mission failure. 
  We propose a hierarchical perception-aware exploration framework for a stereo-equipped unmanned aerial vehicle (UAV) that explicitly incorporates perceptual quality into frontier selection and yaw planning during exploration. Our approach (i) associates each candidate frontier with an expected feature quality using a global feature map, prioritising visually informative subgoals, and (ii) optimises a continuous yaw trajectory along the planned path to maintain stable feature tracks.
  We evaluate our method in simulation across environments with varying texture levels and in real-world indoor experiments with largely textureless walls. Compared to baselines that ignore feature quality and/or omit continuous yaw optimisation, our method achieves more reliable feature tracking, reduced odometry drift, and up to 48\% higher coverage before odometry error exceeds specified thresholds.
\end{abstract}

\begin{IEEEkeywords}
View Planning for SLAM, Motion and Path Planning, Aerial Systems: Perception and Autonomy
\end{IEEEkeywords}

\section{Introduction}
\label{sec:intro}


\IEEEPARstart{A}{utonomous} exploration is a key capability for robots operating in unknown environments. Most existing systems prioritise coverage efficiency by greedily selecting frontiers or maximising information gain \cite{yamauchi1998frontier,zhou_fuel_2020,zhang_falcon_2024,popovic2024learning, polvara2020next}. These strategies typically assume reliable state estimation and treat localisation as a separate module. In practice, however, localisation quality is tightly coupled to sensing conditions and to the information content along the robot’s trajectory.


For unmanned aerial vehicles (UAVs), visual-inertial odometry (VIO) is a common choice for lightweight state estimation~\cite{qin2018vins,geneva2020openvins}. VIO performance depends on the availability of high-quality trackable visual features in the environment. When the exploration policy guides the UAV into feature-sparse regions, visual tracking degrades, odometry drift increases, and the resulting map can become inconsistent or unusable \cite{peng2022rwt, hardt2020monocular}. This reveals a fundamental challenge in autonomous exploration: a robot must explore unknown areas while ensuring it observes enough visual features to maintain a stable state estimate. Prior work on perception-aware navigation mitigates this issue by adapting viewpoint trajectories to improve feature co-visibility and tracking robustness~\cite{zhang_perception-aware_2018,chen_apace_2024,yu_perception-aware_2025}. However, these methods are predominantly designed for point-to-point navigation and often assume partially known environments. They do not address the core exploration problem of trading off frontier progress against perceptual reliability. Consequently, the trade-off between frontier-driven exploration and feature observability within a unified exploration framework remains an open problem.


This paper presents a hierarchical perception-aware autonomous exploration framework that embeds feature quality into both frontier selection and trajectory optimisation. First, we estimate the expected visual feature quality of candidate frontiers using a global feature map, and prioritise subgoals that are both exploratory and visually informative. Second, given a planned position trajectory, we optimise a continuous yaw profile to maintain stable feature tracks by keeping informative regions in view and reducing adverse relative angular motion. Together, these components actively shape where the UAV explores and how it observes, improving odometry robustness with minimal impact on coverage efficiency.
\begin{figure}[!t]
    \centering
    \includegraphics[width=0.48\textwidth]{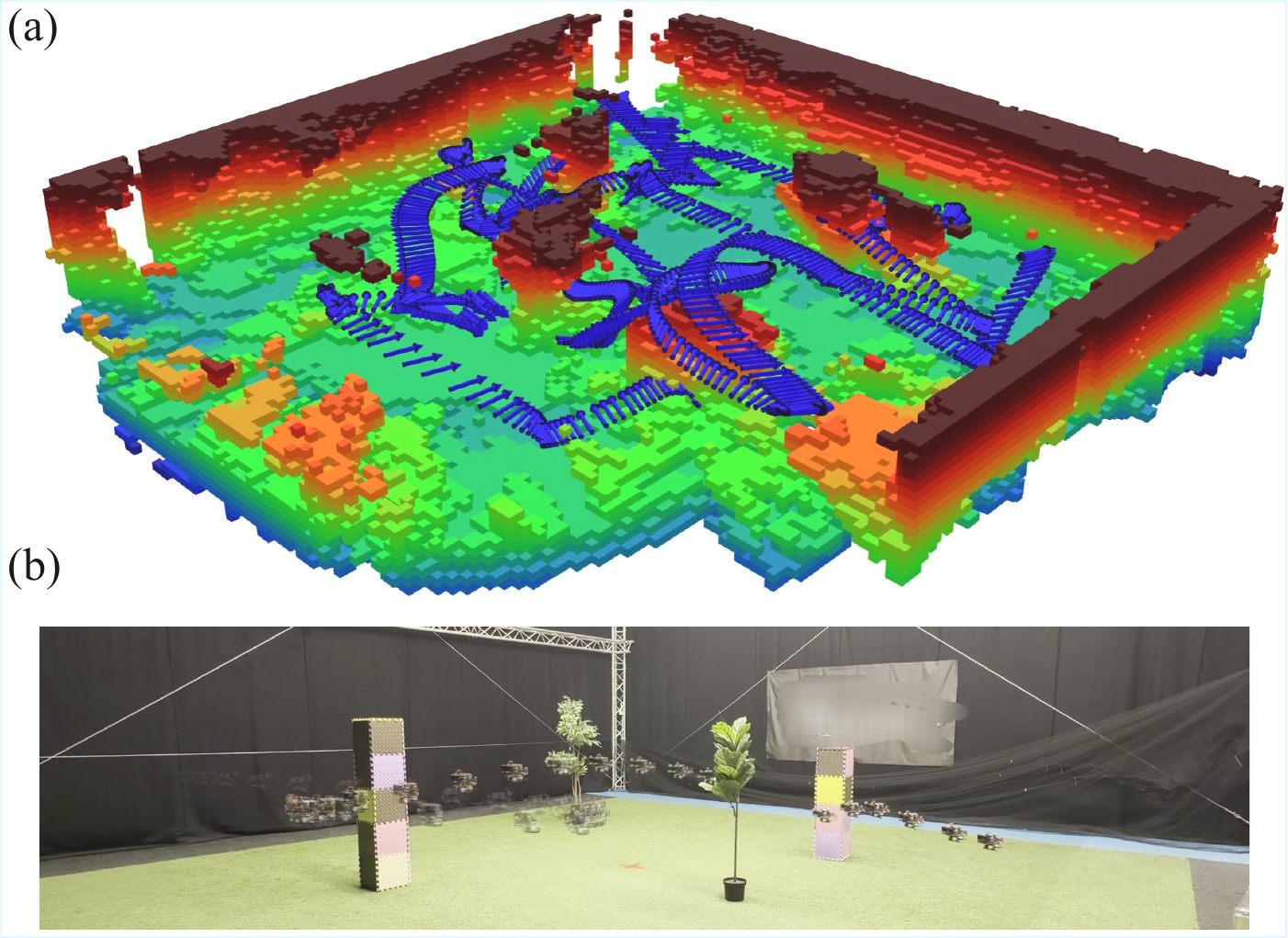}
    \caption{Our approach for perception-aware exploration applied on a UAV. (a) Odometry trajectory during an exploration task (blue arrows); arrow headings indicate the planned yaw angle. (b) Corresponding camera view. Our method maintains reliable feature tracking during exploration.}
    \label{fig:cover}
\end{figure}
We make the following two key claims: 
\begin{enumerate}
    \item Our perception-aware exploration pipeline improves feature tracking compared to perception-agnostic exploration, while maintaining exploration progress.
    \item More reliable feature tracking reduces odometry drift and increases exploration success under realistic VIO error thresholds.
\end{enumerate}

We validate the proposed system in simulation across environments with varying texture levels and in real-world indoor experiments with largely textureless walls. The code implementation can be found at \url{https://github.com/smoggy-P/Perception-Aware-Exploration}.



\section{Related Work}
\label{sec:related}


Our work relates to (i) autonomous exploration methods that select subgoals to maximise coverage or information gain, and (ii) perception-aware planning approaches that explicitly optimise motion to improve vision-based localisation.

\subsection{Autonomous Exploration}
\label{sec:autonomous_exploration}
Autonomous exploration has been studied extensively in robotics. Frontier-based exploration is a classical approach, where the robot explores unknown environments by navigating toward the boundaries between known and unknown space~\cite{yamauchi1998frontier}, which are defined as frontiers.
For UAVs, recent work has focused primarily on improving exploration efficiency.
Several works~\cite{cieslewski_rapid_2017, zhou_fuel_2020, tao2022seer, yuan2024exploring} investigate strategies for selecting the next frontier as a sub-goal.
Beyond purely greedy exploration, several studies propose hierarchical and more global decision-making that mitigates the myopic nature of greedy frontier policies~\cite{cao_tare_2021, zhang_falcon_2024, dong2025eden}.
In addition, learning-based exploration approaches have been proposed to further improve efficiency and robustness in complex environments~\cite{popovic2024learning, cao2025dare, ruckin2022adaptive}. 

Many high-performance exploration systems rely on LiDAR sensing, which provides dense geometric measurements and is inherently less sensitive to visual texture~\cite{zhu2024code, geng2025epic, liu2025flare}, while small UAV platforms usually leverage VIO due to payload and power constraints. In such systems, state estimation quality strongly depends on the availability of trackable visual features: when the robot enters feature-sparse regions, tracking degrades and odometry drift increases, which can corrupt maps and terminate exploration prematurely~\cite{peng2022rwt, hardt2020monocular}. Despite this coupling between exploration and localisation, most exploration frameworks~\cite{zhou_fuel_2020, tao2022seer, yuan2024exploring} still optimise coverage while implicitly assuming sufficiently reliable state estimation. The interaction between exploration strategy and visual localisation robustness thus remains largely underexplored.

\subsection{Perception-aware Navigation}
\label{sec:perception_aware_navigation}
Perception-aware navigation aims to improve localisation accuracy 
by explicitly incorporating perceptual quality into motion planning~\cite{zhang_perception-aware_2018,chen_apace_2024}. 
Unlike exploration, which seeks to maximise coverage of unknown space, 
it typically assumes a predefined goal 
and plans a trajectory that maintains favourable visual conditions 
along the way. 

Existing works can be categorised based on how they model perceptual quality.
Direct methods estimate localisation uncertainty from image measurements and guide sampling-based planning towards visually informative trajectories~\cite{costante_perception-aware_2017}.
Receding-horizon navigation enables reaching a goal while maintaining stable visual feature tracks~\cite{zhang_perception-aware_2018}.
Semantic information has also been leveraged to avoid perceptually degraded regions~\cite{bartolomei_perception-aware_2020, bartolomei_semantic-aware_2021}. More recently, Chen~\textit{et al.}~\cite{chen_apace_2024} introduce a differentiable visibility model that softly penalises feature loss outside the camera field of view and optimises a continuous yaw profile accordingly.
Our work is closely related as we also optimise yaw to maintain feature trackability; however, Chen~\textit{et al.}~\cite{chen_apace_2024} assume that relevant landmarks are known in advance to evaluate visibility, whereas we operate during exploration with an incremental map and uncertain perceptual conditions. Similarly, LA~\cite{yu_perception-aware_2025} optimises feature covisibility and produces a smooth yaw trajectory for point-to-point navigation. In contrast, we extend these ideas to autonomous exploration by coupling yaw planning with feature-aware subgoal selection, and by explicitly accounting for fast motions that can degrade tracking (e.g. via relative angular-velocity considerations), which are not explicitly handled in LA.

Beyond point-to-point navigation, simultaneous localisation and mapping (SLAM)-aware exploration methods explicitly consider the interaction between exploration and localisation uncertainty. Classical active SLAM approaches select between visiting unexplored areas and revisiting mapped regions by reasoning over pose-graph or map uncertainty~\cite{valencia_active_2012,vallve_active_2015}. Loop-aware exploration encourages revisiting previously mapped regions to create useful loop closures~\cite{pittol_loop-aware_2022}. More recent systems introduce explicit graph-stabilisation behaviours: Deshpande~\textit{et al.}~\cite{deshpande_lighthouses_2023} create and revisit visually informative ``lighthouses'' to improve loop closure for low-compute robots with narrow fields of view, while Maheshwari~\textit{et al.}~\cite{maheshwari_region_2025} perform region-based exploration, pose-graph stabilisation, and keyframe marginalisation to improve scalability. These works show that exploration should not be planned independently of SLAM quality. However, they mainly address localisation robustness at the pose-graph or loop-closure level by reasoning about when and where to revisit known areas. Our work addresses a complementary aspect: maintaining front-end visual observability during autonomous exploration. Specifically, we couple feature-aware frontier selection with perception-aware yaw optimisation, enabling the robot to expand into unknown space while preserving trackable visual features.



\section{Our Approach}
\label{sec:method}

\begin{figure}[!t]
    \centering
    \includegraphics[width=0.46\textwidth]{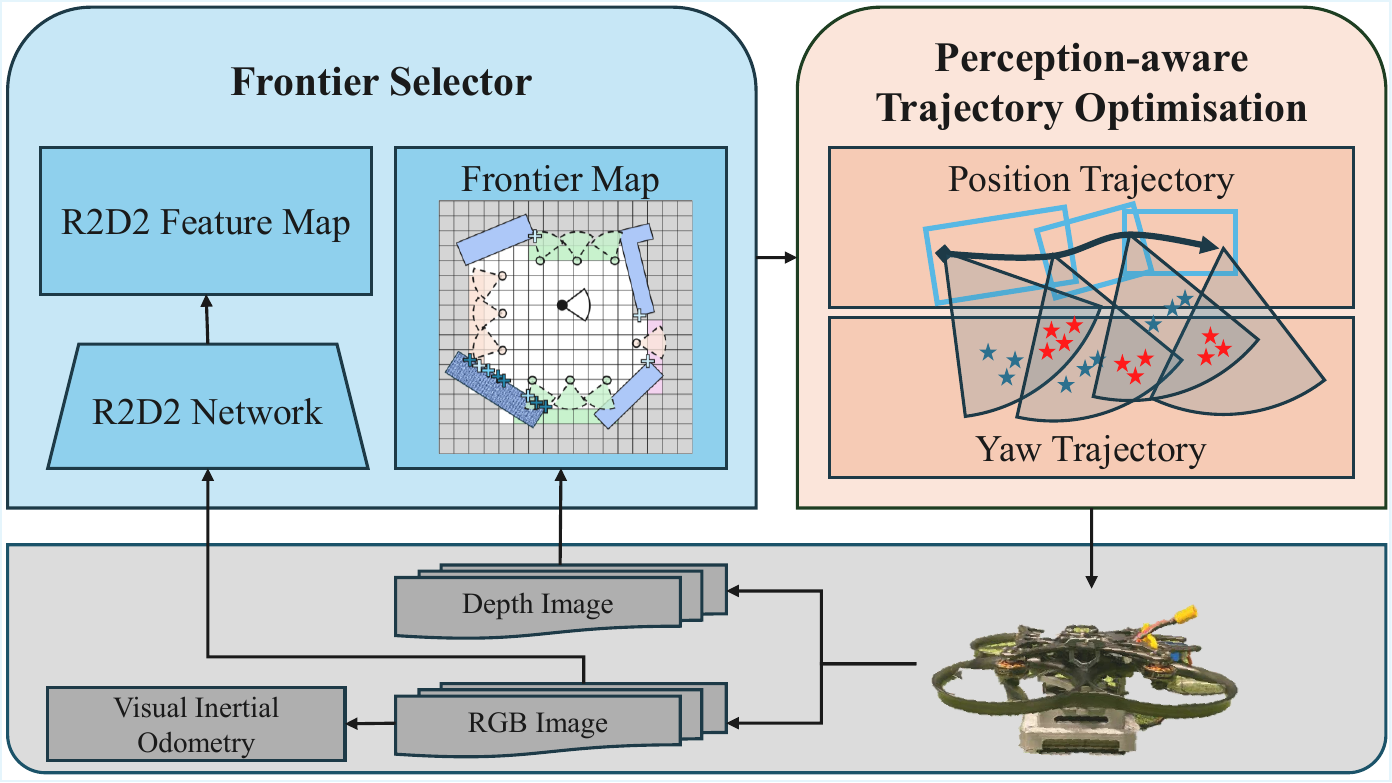}
    \caption{Our perception-aware exploration framework. Two key components are frontier selector and perception-aware trajectory optimisation.}
    \label{fig:system}
\end{figure}

We introduce our perception-aware exploration framework for a UAV equipped with an RGB-D camera, targeting autonomous exploration in feature-limited environments. The goal is to uncover the mapped free space while maintaining reliable visual tracking to avoid odometry drift. Our key contribution is to embed perceptual quality directly into the exploration loop. Fig.~\ref{fig:system} summarises our hierarchical framework. We (i) select the next subgoal among frontier viewpoints by considering exploration utility and expected feature observability (Sec.~\ref{sec:frontier_selector}), and (ii) generate a collision-free trajectory to this subgoal while optimising the yaw profile to support stable feature tracking along the path (Sec.~\ref{sec:traj_opt}). This system enables the UAV to actively prefer frontiers that are both informative and perceptually feasible, and to execute motions that preserve perceptual quality during travel.

\subsection{Frontier Selector}
\label{sec:frontier_selector}

We build on a frontier-based exploration strategy to select the next subgoal during a mission. Following the concept of the Frontier Information Structure (FIS) introduced by Zhou \textit{et al.}~\cite{zhou_fuel_2020}, we cluster frontier voxels and sample a set of candidate viewpoints around each cluster. Each viewpoint candidate is denoted as 
\begin{equation}
    \mathbf{v}_i = [\,\mathbf{p}_i^\top,\, \psi_i\,]^\top,
\end{equation}
where $\mathbf{p}_i = [x_i, y_i, z_i]^\top$ represents the 3D position and $\psi_i$ denotes the UAV yaw angle at that viewpoint.

\begin{figure}[h]
    \centering
    \includegraphics[width=0.45\textwidth]{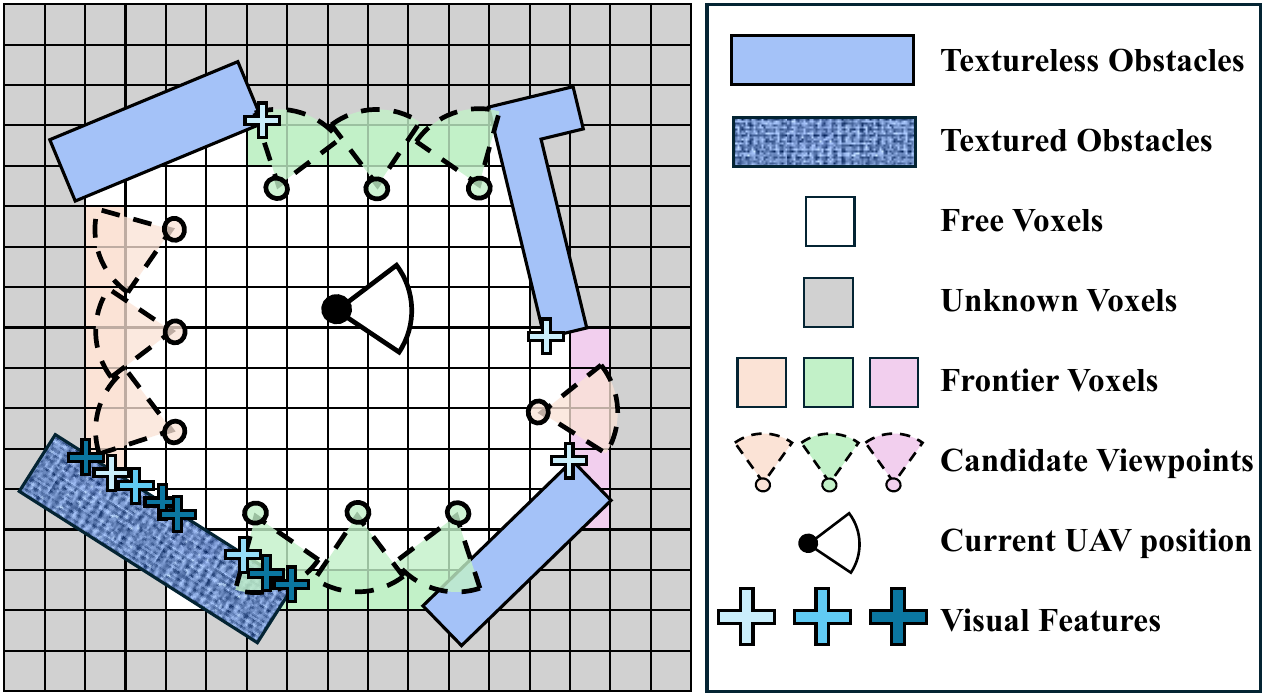}
    \caption{Our frontier selector. Frontier voxels are clustered by proximity, and candidate viewpoints are sampled from known free space to observe the frontiers. Visual features are overlaid, where blue intensity indicates feature quality: textured obstacles yield more and higher-quality features than textureless surfaces. Our selector prefers bottom-left viewpoints, which are farther from the current UAV position but provide stronger feature support than closer viewpoints.}
    \label{fig:FIS}
\end{figure}

A key requirement for perception-aware frontier selection is an estimate of where reliable visual constraints are likely to be available. We therefore maintain a global visual feature map constructed online from the visual front-end, associating spatial locations with a feature quality score:
\begin{equation}
    \mathcal{F} = \{\, (\mathbf{f}_j, s_j) \mid \mathbf{f}_j \in \mathbb{R}^3,\, s_j \in \mathbb{R}^+ \,\},
\end{equation}
where $\mathbf{f}_j$ denotes the 3D position of the $j$-th feature and $s_j$ is its quality score provided by the front-end. This representation is agnostic to the feature type and compatible with both classical and learned detectors. In our experiments, we use R2D2 \cite{revaud2019neurips} as a practical instantiation because it provides a per-feature confidence score that can be directly used as the feature-quality weight. Fig.~\ref{fig:FIS} illustrates the selection process.

Given the current UAV state $(\mathbf{p}_c, \psi_c)$, we score each viewpoint candidate $\mathbf{v}_i$ using a composite cost function that balances three terms: 
the \textit{distance cost}, \textit{coverage utility}, and \textit{visual feature utility}. The total frontier score is defined as
\begin{equation}
  \begin{aligned}
      J(\mathbf{v}_i) = -w_d \|\mathbf{p}_c - \mathbf{p}_i\| 
      &+ w_c Q_c(\mathbf{v}_i) \\
      &+ w_v \tanh\!\bigl(Q_v(\mathbf{v}_i) - \hat{Q}_v\bigr),
  \end{aligned}
  \label{eq:frontier_score}
  \end{equation}
where $w_d$, $w_c$, and $w_v$ are weighting coefficients. 

\noindent \textit{Distance term}. The Euclidean distance $\|\mathbf{p}_c - \mathbf{p}_i\|$ penalises long travel and encourages efficient exploration.

\noindent \textit{Coverage term}. $Q_c(\mathbf{v}_i)$ measures expected exploration progress, computed as the number of unknown voxels that would become observed from $\mathbf{v}_i$. 

\noindent \textit{Visual feature term}. $Q_v(\mathbf{v}_i)$ provides a lightweight empirical proxy for local perceptual support by aggregating the quality scores of visible features:
\begin{equation}
    Q_v(\mathbf{v}_i) = \sum_{\mathbf{f}_j \in \mathcal{V}(\mathbf{p}_i, \psi_i)} s_j,
    \label{eq:feature_utility}
\end{equation}
where $\mathcal{V}(\mathbf{p}_i, \psi_i) \subseteq \mathcal{F}$ is the subset of features visible from viewpoint $\mathbf{v}_i$. Since empirical results shows that additional features beyond a certain threshold tend to provide diminishing returns for this high-level decision, we map $Q_v$ to a bounded score via the $\tanh$ function and the hyperparameter $\hat{Q}_v$, so that the score grows approximately linearly when $Q_v\ll \hat{Q}_v$ and saturates when $Q_v\gtrsim \hat{Q}_v$. This encourages the selector to prefer frontiers that remain perceptually supported, rather than purely maximising frontier progress.

Finally, the next exploration subgoal $\left[\:\mathbf{p}_g^\top,\, \psi_g\:\right]$ is chosen as the highest-scoring candidate: 
\begin{equation}
  \left[\:\mathbf{p}_g^\top,\, \psi_g\:\right]^\top = \arg \max_{\mathbf{v}_i \in \mathcal{V}} J(\mathbf{v}_i),
\end{equation}
where $\mathcal{V}$ is the viewpoint candidate set. The selected subgoal balances exploration progress with expected feature support and serves as input to our yaw-aware trajectory optimisation.

\subsection{Perception-Aware Trajectory Generation}
\label{sec:traj_opt}

Once the frontier selector returns the next exploration subgoal $\left(\:\mathbf{p}_g^\top,\, \psi_g\:\right)$, we generate a collision-free trajectory that is also perceptually reliable. To exploit the differential flatness of quadrotor dynamics \cite{mellinger2011minimum}, we decompose the planning problem into two coupled parts:
(i) a position trajectory $\mathbf{p}(t)$ that guarantees safety and dynamic feasibility, and (ii) a yaw trajectory $\psi(t)$ that is optimised to maintain visual observability along the motion.

\subsubsection{Position Trajectory Optimisation}

We follow a two-stage procedure similar to Chen \textit{et al.}~\cite{chen2022rast}.
First, a Rapidly exploring Random Tree (RRT)-based planner provides a collision-free reference path connecting the current state to the selected subgoal. 
Second, we refine this path into a smooth, dynamically feasible trajectory using minimum-snap optimisation with Bézier curves. The position trajectory $\mathbf{p}(t)$ is composed of $M$ Bézier segments, each corresponding to a safe flight corridor \cite{liu2017planning}. 
Let $\mathbf{p}_j(t)$ denote the $j$-th Bézier segment defined over $t \in [t_{j-1}, t_j]$:
\begin{equation}
    \mathbf{p}_j(t) = \sum_{k=0}^{N} \mathbf{p}_{j,k} \, B_k^N(\tau), \quad \tau = \frac{t - t_{j-1}}{t_j - t_{j-1}},
    \label{eq:bezier_segment}
\end{equation}
where $\mathbf{p}_{j,k} \in \mathbb{R}^3$ are the control points of the Bézier curve, 
$B_k^N(\tau) = \binom{N}{k}(1-\tau)^{N-k}\tau^k$ are the Bernstein basis polynomials of order $N$. The optimisation objective minimises the overall snap:
\begin{equation}
  \begin{aligned}
      \min_{\{\mathbf{p}_{j,k}\}} \quad 
      & \sum_{j=1}^{M} 
      \int_{t_{j-1}}^{t_j} 
      \left\| \frac{d^4 \mathbf{p}_j(t)}{dt^4} \right\|^2 dt \\
      \text{s.t.} \quad
      & \mathbf{p}_{1,0} = \mathbf{p}_c, \quad \mathbf{p}_{M,N} = \mathbf{p}_g, \\
      & \mathbf{p}_{j,N} = \mathbf{p}_{j+1,0}, \quad \forall j=1,2,\ldots,M-1, \\
      & \|\dot{\mathbf{p}}_j(t)\| \leq v_{\max}, \quad 
                    \|\ddot{\mathbf{p}}_j(t)\| \leq a_{\max},\\
      & \mathbf{p}_{j,k} \in \hat{\mathcal{E}}_j, \quad \forall j=1,2,\ldots,M, \forall k=0,1,\ldots,N.
  \end{aligned}
  \label{eq:min_snap_full}
  \end{equation}
Here, $\mathbf{p}_c$ and $\mathbf{p}_g$ denote the current and goal positions, 
and $\hat{\mathcal{E}}_j$ is the convex safe corridor of the $j$-th segment. $v_{\max}$ and $a_{\max}$ are the maximum velocity and acceleration of the UAV.
The four constraints correspond to the boundary conditions, continuity constraints, dynamic limits, and collision-free constraints, respectively.

\subsubsection{Yaw Trajectory Optimisation}
\label{sec:yaw_opt}
\begin{figure}[h]
    \centering
    \includegraphics[width=0.3\textwidth]{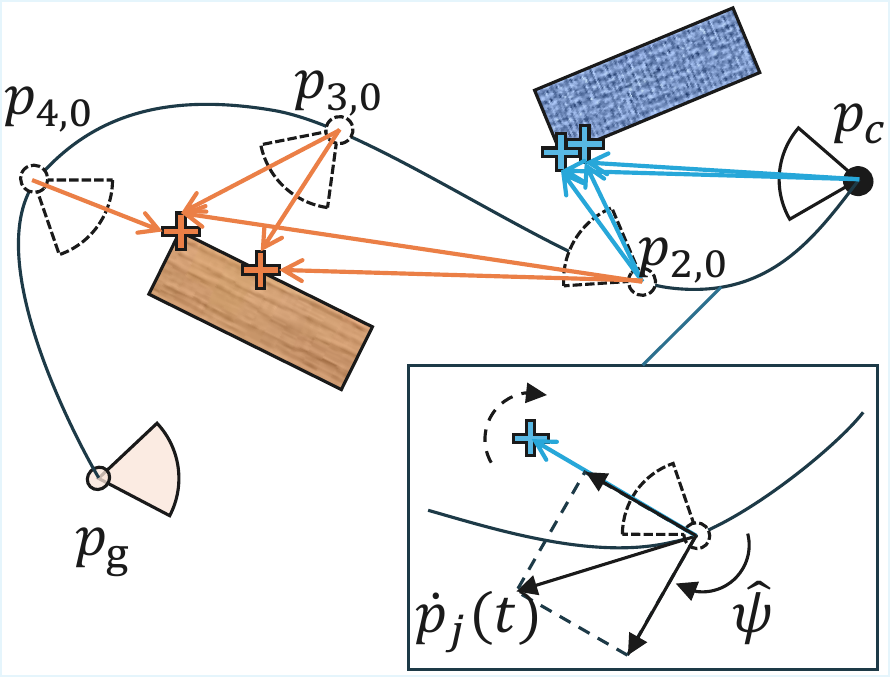}
    \caption{Our yaw angle trajectory optimisation procedure. Dashed field of view sectors represent the waypoint yaws $\{\psi_j\}$ and arrows with different colours show covisible feature sets $\{\mathcal{F}_j\}$. The zoomed-in view shows the calculation of the relative angular velocity. For any point on trajectory, we decompose the velocity $\dot{\mathbf{p}_j}(t)$ in the direction of the feature vectors $\mathbf{f}_i - \mathbf{p}_j(t)$ and the perpendicular direction, expecting the yaw rate to compensate the relative angular velocity.}
    \label{fig:yaw_opt}
\end{figure}

While $\mathbf{p(t)}$ ensures safe motion, it does not guarantee that the camera maintains sufficient visual constraints throughout the flight. We therefore optimise the yaw trajectory $\psi(t)$ in two steps: a covisibility-based waypoint sampling, followed by a continuous yaw refinement that reduces feature motion in the image. Our sampling algorithm is described in Algorithm~\ref{alg:covisibility_sampling}. Covisibility sampling yields a set of waypoint yaw angles $\{\psi_j\}$ and corresponding covisible feature sets $\{\mathcal{F}_j\}$, as shown in Fig.~\ref{fig:yaw_opt}. In Algorithm~\ref{alg:covisibility_sampling}, \textsc{GetVisibleFeatures}$(\mathbf{p},\psi)$ returns the set of map features that fall within the camera field of view at viewpoint $(\mathbf{p},\psi)$ and are not occluded by obstacles; we evaluate visibility via ray-casting (line-of-sight checks) in the occupancy map. The threshold $\tau_{\text{cov}}$ specifies the minimum aggregate feature quality that must be shared between consecutive waypoints. The fallback condition $k > 10$ limits the number of yaw sampling attempts per waypoint: if no yaw achieves sufficient covisibility within $k$ times of search, we conclude that the local environment is feature-poor and linearly interpolate the yaw toward the goal. This avoids excessive computation in textureless corridors while still guaranteeing that the algorithm terminates.

\begin{algorithm}[t]
  \caption{Covisibility Sampling}
  \label{alg:covisibility_sampling}
  \begin{algorithmic}[1]
  \REQUIRE Control points $\{\mathbf{p}_{j, k}\}$, time allocations $\{t_j\}$, initial pose $(\mathbf{p}_{c}, \psi_{c})$, target yaw $\psi_{g}$, covisibility threshold $\tau_{\text{cov}}$, yaw sampling step $\Delta \psi$
  \ENSURE Sampled waypoint yaws $\{\psi_j\}$, covisible feature sets $\{\mathcal{F}_j\}$
  \STATE $\psi_{\text{last}} \leftarrow \psi_{c}$
  \FOR{$j = 1$ to $M-1$}
      \STATE $\mathcal{V}_{\text{last}} \leftarrow \textsc{GetVisibleFeatures}(\mathbf{p}_{j,0}, \psi_{\text{last}})$
      \STATE $k \leftarrow 0$
      \STATE $\psi_s \leftarrow \psi_{\text{last}}$
      \WHILE{true}
          \STATE $\mathcal{V}_s \leftarrow \textsc{GetVisibleFeatures}(\mathbf{p}_{j+1,0}, \psi_s)$
          \STATE $\mathcal{V}_{\text{rep}} \leftarrow \mathcal{V}_s \cap \mathcal{V}_{\text{last}}$ \COMMENT{covisible features}
          \STATE $w_{\text{cov}} \leftarrow \sum_{\mathbf{f}_i \in \mathcal{V}_{\text{rep}}} s_i$ \COMMENT{sum of feature scores}
          \IF{$w_{\text{cov}} > \tau_{\text{cov}}$}
              \STATE $\mathcal{F}_j \leftarrow \mathcal{V}_{\text{rep}}$, $\psi_j \leftarrow \psi_s$, $\psi_{\text{last}} \leftarrow \psi_s$
              \STATE \textbf{break}
          \ELSE
              \IF{$\psi_{g} > \psi_{\text{last}}$}
                  \STATE $\psi_s \leftarrow \psi_s - \Delta \psi$ 
              \ELSE
                  \STATE $\psi_s \leftarrow \psi_s + \Delta \psi$
              \ENDIF \COMMENT{sample yaw angle toward the goal yaw angle}
              \STATE $k \leftarrow k + 1$
              \IF{$k > 10$}
                  \STATE $\psi_j \leftarrow \psi_{\text{last}} + (\psi_{g} - \psi_{\text{last}}) \times \frac{j}{M}$ \COMMENT{if no covisibility is found, sample yaw angle toward the goal yaw angle linearly}
                  \STATE $\mathcal{F}_j \leftarrow \emptyset$
                  \STATE $\psi_{\text{last}} \leftarrow \psi_j$
                  \STATE \textbf{break}
              \ENDIF
          \ENDIF
      \ENDWHILE
  \ENDFOR
  \STATE Append final yaw: $\psi_{M} \leftarrow \psi_{g}$
  \end{algorithmic}
  \end{algorithm}

Covisibility at waypoints alone does not prevent visual degradation between waypoints, especially when features undergo high relative angular motion during translation. Hence, we further optimise the continuous yaw trajectory to maintain feature visibility and reduce the relative angular motion. The yaw angle trajectory $\psi(t)$ is represented as a 1D Bézier curve parameterised by control points:
\begin{equation}
    \label{eq:bezier_curve_psi}
    \psi_j(t) = \sum_{k=0}^{n} \psi_{j,k} B_k^n(\tau), \quad \tau = \frac{t - t_{j-1}}{t_j - t_{j-1}},
\end{equation}
where $B_k^n(\tau)$ are Bernstein basis polynomials and $\psi_{j,k}$ are the control points. The problem is formulated as:
\begin{equation}
    \min_{\{\psi_{j,k}\}} \; \sum_{j=1}^{M} J_{\text{percept}}(\psi_j(t)) + \lambda_\psi J_{\text{smooth}}(\psi_j(t)),
\end{equation}
where $J_{\text{percept}}$ calculates the integrated relative angular velocity cost over the trajectory, $J_{\text{smooth}}$ regularises the yaw rate for smooth motion, and $\lambda_\psi$ is a weighting coefficient.

For a given point $\mathbf{p}_j(t)$ on the trajectory, the target visual features are $\mathbf{f}_i \in \mathcal{F}_j$. As the UAV translates, each feature's bearing angle in the horizontal plane changes. We define the desired yaw rate $\hat{\psi}_{j,i}(t)$ as the rate at which the bearing to feature $\mathbf{f}_i$ changes due to the UAV's translational velocity $\dot{\mathbf{p}}_j(t)$. Let $\mathbf{d}_{j,i}(t) = \mathbf{f}_i - \mathbf{p}_j(t)$ be the displacement vector; the bearing rate is the $z$-component of the angular velocity induced by the cross product:
\begin{equation}
    \hat{\psi}_{j,i}(t) = \left(\frac{\mathbf{d}_{j,i}(t)}{\|\mathbf{d}_{j,i}(t)\|} \times \dot{\mathbf{p}}_j(t)\right)_z \cdot \frac{1}{\|\mathbf{d}_{j,i}(t)\|},
    \label{eq:desired_yaw_rate}
\end{equation}
where $(\cdot)_z$ extracts the $z$-component (i.e. the component about the vertical axis). Intuitively, $\hat{\psi}_{j,i}(t)$ represents how fast the UAV should rotate in yaw to keep feature $\mathbf{f}_i$ stationary in the camera frame.

We wish the actual yaw rate $\dot{\psi}_j(t)$ to track the feature-weighted average of these desired rates. The perceptual cost is:
\begin{equation}
    J_{\text{percept}}(\psi_j(t)) = \int_{t_{j-1}}^{t_j} \sum_{\mathbf{f}_i \in \mathcal{F}_j} s_i \left( \hat{\psi}_{j,i}(t) - \dot{\psi}_j(t) \right)^2 dt,
\end{equation}
The smoothness term regularises the yaw rate:
\begin{equation}
    J_{\text{smooth}}(\psi_j(t)) = \int_{t_{j-1}}^{t_j} \left( \dot{\psi}_j(t) \right)^2 dt,
\end{equation}
The cost function is smooth and differentiable with respect to the control points $\psi_{j,k}$:
    \begin{align}
    \frac{\partial J_{\text{percept}}(\psi_j(t))}{\partial \psi_{j,k}}
    &=\int_{t_{j-1}}^{t_j}\sum_{i}2s_i\big(\dot\psi_j(t)-\hat\psi_{j,i}(t)\big)
    \frac{\partial \dot\psi_j(t)}{\partial \psi_{j,k}}\,dt,
    \end{align}
where $\frac{\partial \dot{\psi}_j(t)}{\partial \psi_{j,k}}$ is derived from the Bézier parameterisation in Eq.~\eqref{eq:bezier_curve_psi}. The yaw trajectory should satisfy kinematic limits:
\begin{equation}
    |\dot{\psi}_j(t)| \leq \dot{\psi}_{\max}, \quad |\ddot{\psi}_j(t)| \leq \ddot{\psi}_{\max}, \quad \forall\, t \in [t_{j-1}, t_j],
    \label{eq:yaw_constraints}
\end{equation}
where $\dot{\psi}_{\max} = 1.5$\,rad/s and $\ddot{\psi}_{\max} = 3.0$\,rad/s$^2$. We solve the constrained optimisation with sequential quadratic programming (SQP) from the NLopt package~\cite{johnson2007nlopt,kraft1994tomp}. 

Crucially, this module contributes a continuous, feature-aware yaw optimisation that improves perceptual quality along the motion, yielding a perception-aware trajectory $\mathbf{p}(t), \psi(t)$ that is both safe and VIO-robust.

\section{Experimental Results}
\label{sec:exp}

%

%
We present experiments designed to assess the effectiveness of our proposed perception-aware exploration framework and support our key claims. In particular, the experiments demonstrate that our pipeline: (i) maintains more reliable visual feature tracking during exploration progress, and (ii) improves feature tracking, leading to higher exploration success rates under realistic VIO error conditions.

\subsection{Experimental Setup}



\begin{figure}[t]
    \centering
    \includegraphics[width=0.45\textwidth]{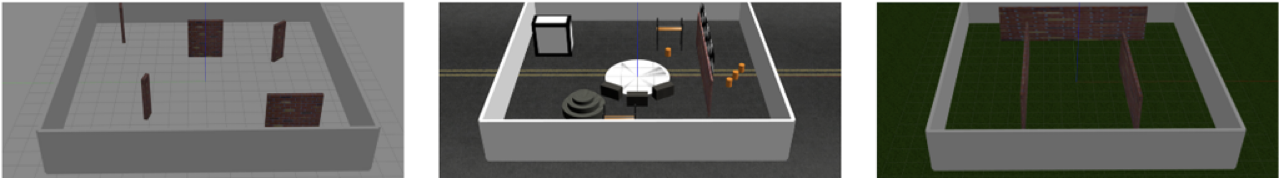}
    \caption{Simulation environments with controlled texture richness. Left to right: \textbf{Low-Texture} (few textured pillars), \textbf{Medium-Texture} (partially textured construction site), and \textbf{High-Texture} (richly textured walls), each $10\,\mathrm{m}\times 10\,\mathrm{m}$.}
    \label{fig:sim_env}
\end{figure}

We evaluate our framework in both simulation and real-world settings. All simulations are conducted in Gazebo. We use depth measurements from a simulated RGB-D camera for occupancy mapping and frontier extraction, and RGB images for feature extraction and VIO. We design three environments with controlled levels of visual texture, as shown in Fig. \ref{fig:sim_env}.

We compare our approach against four baselines. First, FUEL \cite{zhou_fuel_2020} serves as a frontier-based exploration method that prioritises coverage efficiency without considering perceptual quality. Second, we use LA \cite{yu_perception-aware_2025} as a representative perception-aware point-to-point navigation method and couple it with a simple frontier selector to form an exploration pipeline. Together, these baselines represent complementary paradigms: coverage-driven exploration that is agnostic to state estimation quality (FUEL), and perception-aware planning designed for point-to-point navigation rather than full exploration (LA). To enable a fair comparison under different front-end choices, we additionally include two hybrid baselines obtained by replacing the frontier-selection module in FUEL and LA with our perception-aware (PA) frontier selector. In our approach, we build the global feature map $\mathcal{F}$ online using R2D2 features~\cite{revaud2019neurips}, whose confidence scores serve as the feature quality weight $s_j$. For state estimation, we run OpenVINS~\cite{geneva2020openvins} with its default tracking front-end. The feature map $\mathcal{F}$ is used only for frontier scoring and yaw planning and does not replace the VIO front-end.

Each method is run 10 times per environment. Each run ends when the exploration rate (ratio of observed-to-total free voxels) exceeds $95\%$ or the elapsed time exceeds $300$\,s. We define successful exploration as reaching $\geq 95\%$ exploration rate while maintaining VIO position error below a given threshold. This definition captures the practical requirement of completing exploration without localisation failure. We further validate our method through real-world experiments. To facilitate sim-to-real transfer, we employ the Agilicious pipeline \cite{foehn_agilicious_2022} to track generated trajectories in both simulation and on hardware.

\subsection{Feature Tracking Comparison}
\label{sec:feature_tracking_results}
\begin{figure}[t]
    \centering
    \includegraphics[width=0.49\textwidth]{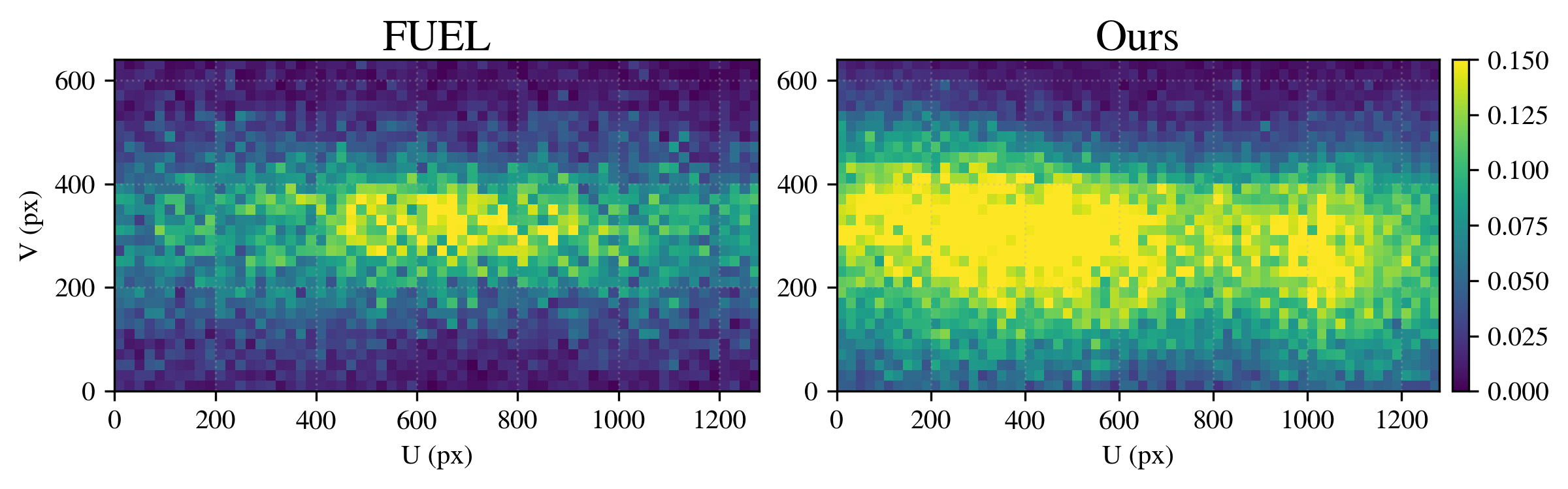}
    \caption{Frame-averaged spatial distribution of tracked visual features in the image plane during exploration. The axes show the pixel coordinates $(u,v)$. For each frame, we build a 2D histogram of tracked feature locations and then average the histogram over time across the entire exploration run; colour indicates the normalised mean feature count per bin.}
    \label{fig:heatmap}
\end{figure}

To support claim (i), we compare all the methods in terms of feature tracking ability during exploration. Fig.~\ref{fig:heatmap} reports the spatial distribution of tracked features in the image plane. For each frame, we accumulate a 2D histogram of tracked feature pixel coordinates and then average it over time; brighter regions therefore indicate image locations that consistently contain more tracked features throughout a run. 

Our method shows higher intensities over a large portion of the image, indicating that it tracks more visual features per frame than FUEL. This is expected since FUEL primarily optimises for coverage efficiency, and thus has no incentive to maintain previously observed features in view when doing so conflicts with rapidly advancing toward unknown space.

\begin{figure}[t]
    \centering
    \includegraphics[width=0.49\textwidth]{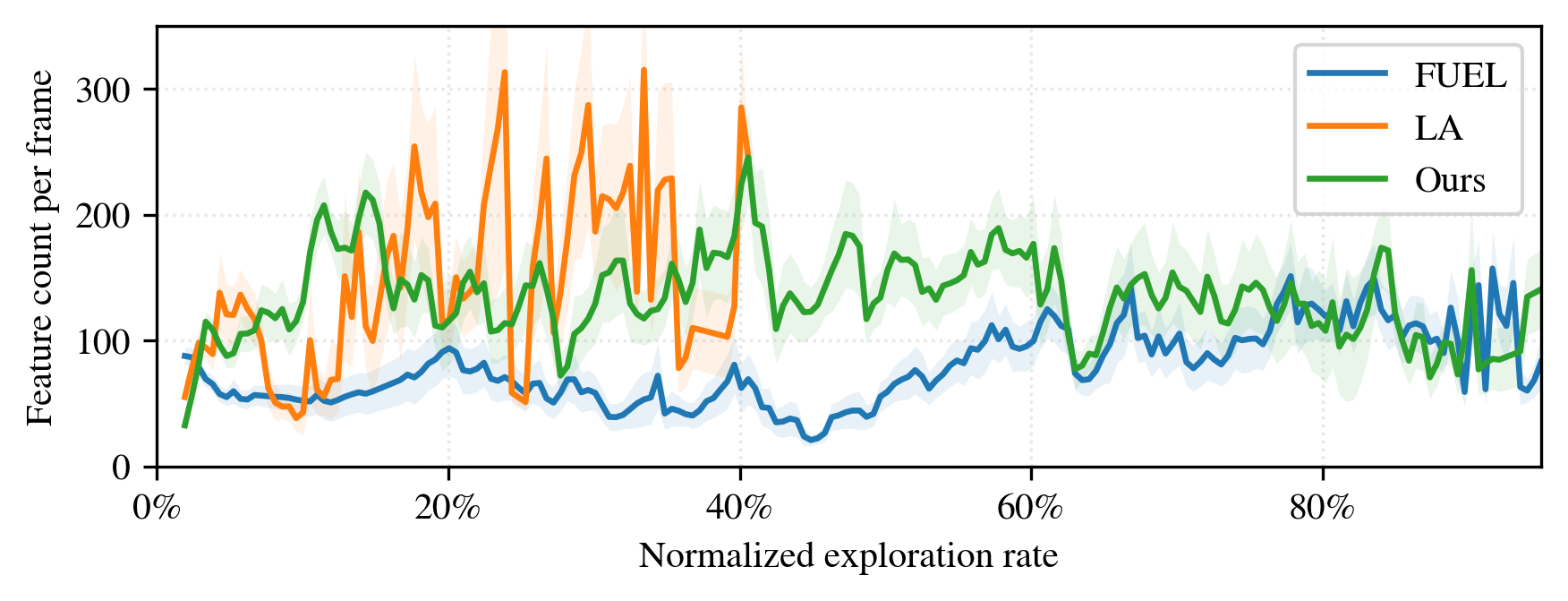}
    \caption{Number of tracked visual features per frame during exploration. The $x$-axis shows normalised exploration progress, defined as the explored free-space volume divided by the total explored volume at run completion. The $y$-axis reports the number of features successfully tracked per frame. Solid lines show the mean over $10$ runs; shaded areas indicate one standard deviation.
}
    \label{fig:feature_exploration}
\end{figure}

Fig.~\ref{fig:feature_exploration} further shows the number of tracked features per frame as exploration progresses. Our method maintains consistently higher feature counts, particularly in the early stages of exploration, which indicates that it actively prioritises frontiers in visually informative regions while still expanding into unknown space. The LA achieves strong feature tracking due to hard constraints on visible features, but it fails to complete the exploration task; this is reflected by its curve terminating at approximately 40\% completion. In contrast, our framework preserves high feature trackability without sacrificing exploration progress, which is critical for maintaining VIO robustness throughout a full mission.

\subsection{Exploration Success Rate Under Odometry Constraints}
\label{sec:exploration_success_rate_results}

To support claim (ii), we compare all methods in terms of exploration success rate under three odometry error thresholds ($1.0\,\mathrm{m}$, $2.0\,\mathrm{m}$, $3.0\,\mathrm{m}$). Table~\ref{tab:ser_thresholds_texture} shows the results across environments. Our method achieves the highest success rate in every environment and for all thresholds. While the Low-Texture map is challenging for all methods, our approach consistently covers more area(up to 48\%) before violating the odometry error threshold. The largest gain in Low-Texture environments and under stricter error thresholds also highlights that our method is especially effective in feature-limited areas. FUEL + PA provides a modest improvement over FUEL, showing that stronger high-level subgoal selection can already improve robustness, while LA + PA underperforms mainly because frequent replanning interacts poorly with LA’s hard-constraint state machine, which can repeatedly re-enter its failure-handling branch.

\begin{table*}[t]
\centering
\caption{Success exploration rate (expressed as mean$\pm$ std dev) under different odometry error thresholds ($1.0\,\mathrm{m}$, $2.0\,\mathrm{m}$, $3.0\,\mathrm{m}$) on environments ordered by texture richness (High $\rightarrow$ Low). Best per column in \textbf{bold}.}
\label{tab:ser_thresholds_texture}
\setlength{\tabcolsep}{3pt}
\begin{tabular}{lccccccccc}
\toprule
\multirow{2}{*}{Method} & \multicolumn{3}{c}{High-Texture Map Exploration Rate} & \multicolumn{3}{c}{Medium-Texture Map Exploration Rate} & \multicolumn{3}{c}{Low-Texture Map Exploration Rate} \\
\cmidrule(lr){2-4} \cmidrule(lr){5-7} \cmidrule(lr){8-10}
& 1.0 m & 2.0 m & 3.0 m & 1.0 m & 2.0 m & 3.0 m & 1.0 m & 2.0 m & 3.0 m \\
\midrule
Ours & \textbf{0.913}{$\pm$0.065} & \textbf{0.913}{$\pm$0.065} & \textbf{0.924}{$\pm$0.049} & \textbf{0.888}{$\pm$0.070} & \textbf{0.928}{$\pm$0.044} & \textbf{0.928}{$\pm$0.044} & \textbf{0.326}{$\pm$0.169} & \textbf{0.427}{$\pm$0.126} & \textbf{0.470}{$\pm$0.139} \\
FUEL & 0.728{$\pm$0.291} & 0.846{$\pm$0.207} & 0.868{$\pm$0.182} & 0.651{$\pm$0.311} & 0.731{$\pm$0.257} & 0.802{$\pm$0.175} & 0.224{$\pm$0.078} & 0.271{$\pm$0.100} & 0.317{$\pm$0.128} \\
FUEL + PA Frontier & 0.815{$\pm$0.085} & 0.828{$\pm$0.058} & 0.828{$\pm$0.058} & 0.667{$\pm$0.147} & 0.771{$\pm$0.093} & 0.771{$\pm$0.093} & 0.321{$\pm$0.202} & 0.337{$\pm$0.206} & 0.325{$\pm$0.203} \\
LA & 0.234{$\pm$0.094} & 0.235{$\pm$0.093} & 0.235{$\pm$0.093} & 0.202{$\pm$0.093} & 0.220{$\pm$0.082} & 0.239{$\pm$0.093} & 0.142{$\pm$0.064} & 0.162{$\pm$0.071} & 0.176{$\pm$0.087} \\
LA + PA Frontier & 0.023{$\pm$0.005} & 0.023{$\pm$0.005} & 0.023{$\pm$0.005} & 0.048{$\pm$0.020} & 0.048{$\pm$0.020} & 0.048{$\pm$0.020} & 0.052{$\pm$0.033} & 0.053{$\pm$0.034} & 0.054{$\pm$0.035} \\
\bottomrule
\end{tabular}
\end{table*}

\begin{figure}[t]
    \centering
    \includegraphics[width=0.45\textwidth]{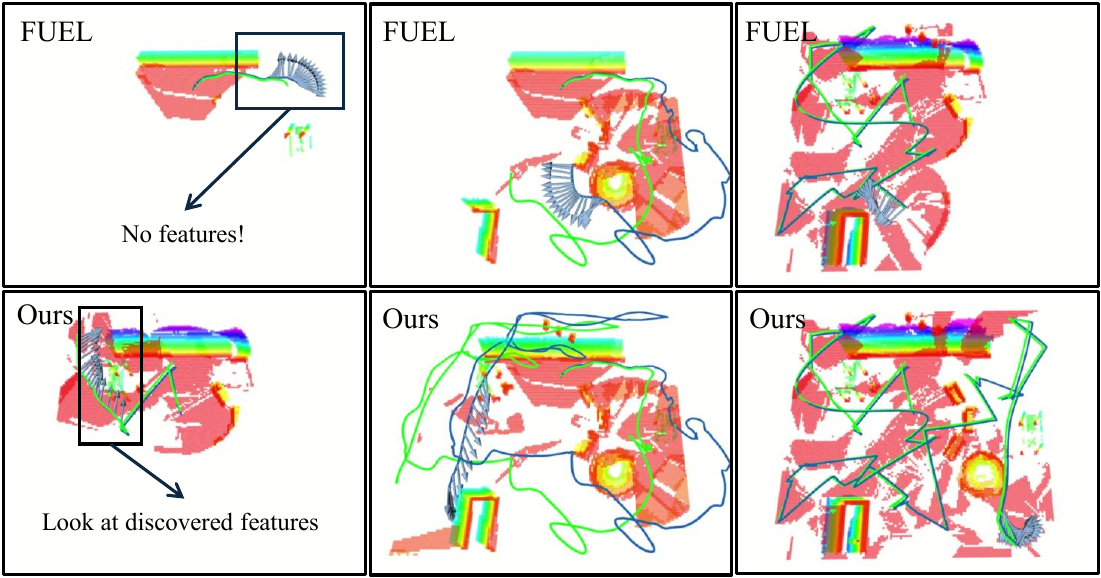}
    \caption{Exploration process in Medium-Texture map. Blue and green lines indicate the odometry and ground-truth pose. Blue arrows indicate yaw angles. Our perception-aware approach enables looking at known features (bottom-left).}
    \label{fig:simulation_exploration}
\end{figure}

Fig.~\ref{fig:simulation_exploration} provides a qualitative view of this effect. Different from FUEL, our method prioritises feature-rich regions by considering perceptual quality in both the frontier selector and trajectory optimisation, resulting in more complete exploration under the same error constraints.

\subsection{Ablation Studies}
We conduct ablation studies to isolate the impact of the key components of our framework: (i) perception-aware frontier selection and (ii) yaw trajectory optimisation. Following Sec.~\ref{sec:feature_tracking_results} and Sec.~\ref{sec:exploration_success_rate_results}, we evaluate feature tracking ability and exploration success rate against the two baselines.

\begin{figure}[h]
    \centering
    \includegraphics[width=0.49\textwidth]{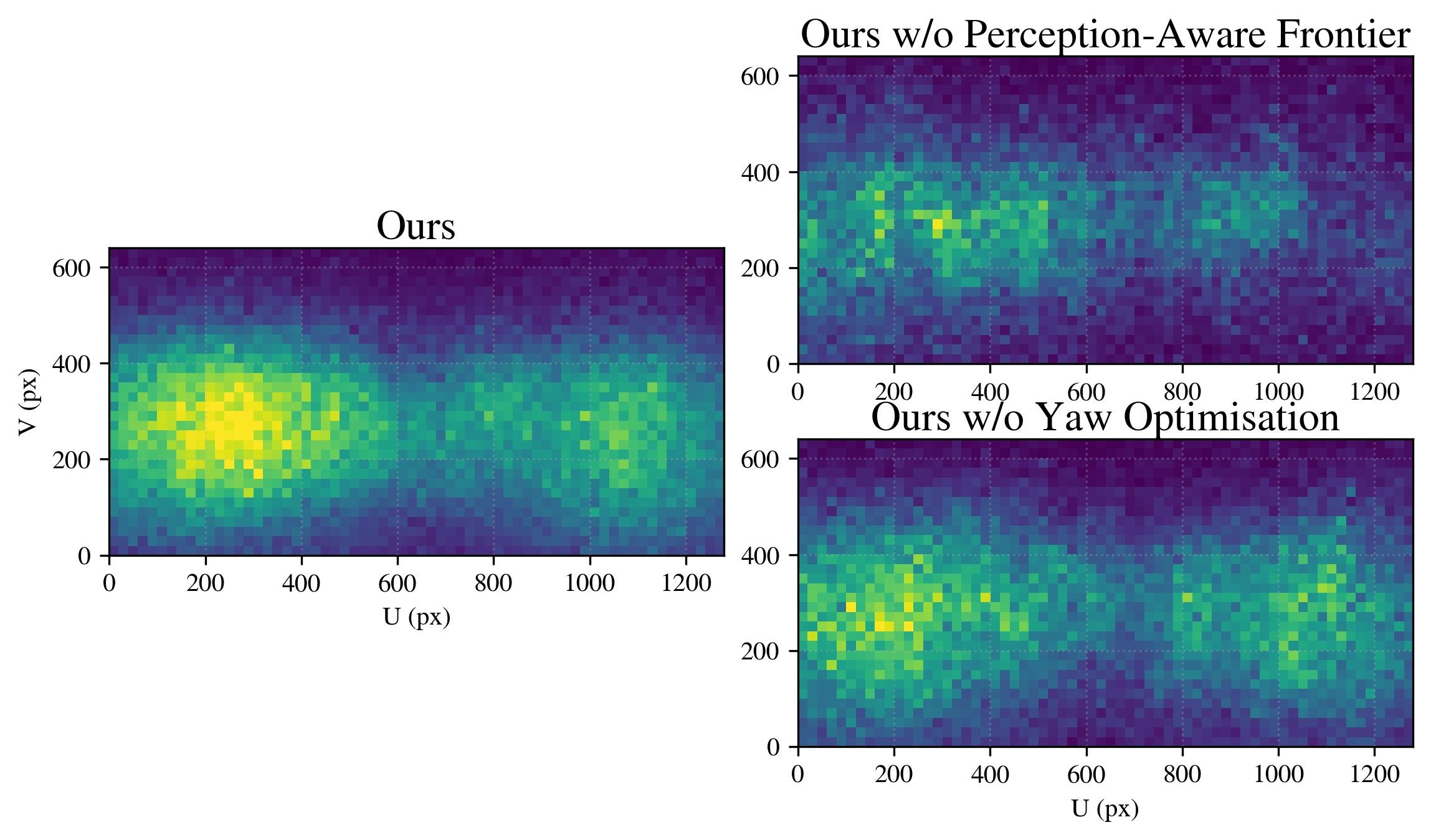}
    \caption{Feature tracking comparison for ablation study. Brighter colours indicate a higher number of tracked features in the image plane. Removing either key component reduces feature intensity, indicating degraded feature tracking.}
    \label{fig:ablation_feature}
\end{figure}

Fig.~\ref{fig:ablation_feature} reports the feature tracking behaviour for the full pipeline and its ablated variants. Removing either component markedly degrades performance, leading to low feature density. Interestingly, a bimodal per-frame feature-density distribution appears in both the full pipeline and the variant without yaw trajectory optimisation. This effect is driven by the perception-aware frontier selector, which balances feature observability with exploration efficiency. By favouring frontiers near texture-rich regions, the camera captures both informative features and unknown space, sustaining reliable tracking without blocking exploration process.

Table~\ref{tab:ablation_ser} reports the exploration success rates on the Low-Texture environment. Removing the perception-aware frontier module causes the largest drop across all thresholds ($54.2\%$ compared to $11.9\%$ drop brought by removing yaw trajectory optimisation), confirming that feature-informed subgoal selection is the dominant contributor to robustness in feature-limited settings. Yaw trajectory optimisation provides additional gains, indicating that continuously shaping the camera motion further improves tracking stability beyond what can be achieved by subgoal choice alone.

\begin{table}[h]
\centering
\footnotesize
\caption{Exploration success rate comparison for ablation study under different odometry error thresholds. Best in \textbf{bold}.}
\label{tab:ablation_ser}
\setlength{\tabcolsep}{3pt}
\begin{tabular}{lccc}
\toprule
\multirow{2}{*}{Method} & \multicolumn{3}{c}{Low-Texture Map Exploration Rate} \\
\cmidrule(lr){2-4}
& 1.0 m & 2.0 m & 3.0 m \\
\midrule
Ours & \textbf{0.326}$\pm$0.169 & \textbf{0.427}$\pm$0.126 & \textbf{0.470}$\pm$0.139 \\
Ours w/o PA Frontier & 0.162$\pm$0.123 & 0.185$\pm$0.131 & 0.209$\pm$0.130 \\
Ours w/o Yaw Optim. & 0.300$\pm$0.057 & 0.368$\pm$0.047 & 0.404$\pm$0.059 \\
\bottomrule
\end{tabular}
\end{table}


\subsection{Real-World Experiments}

We also validate our method in real-world conditions with a custom-built UAV (Fig.~\ref{fig:system}) equipped with a ZED Mini camera providing RGB images and depth measurements and an NVIDIA Jetson Orin NX. Ground-truth poses are obtained from an OptiTrack motion-capture system, and we calculate the odometry error as the Euclidean distance between the OpenVINS position estimate and the corresponding OptiTrack ground-truth position. The testing environment is a $10\,\mathrm{m}\times 10\,\mathrm{m}$ room with randomly distributed obstacles, including artificial plants and stacks of textured cubes. The walls are mostly black and featureless. We performed three exploration trials with each of two layouts, denoted \emph{Sparse Layout} and \emph{Dense Layout}(with less and more textured obstacles). The results are reported in Tab.~\ref{tab:real_world_summary}. Across all six trials, the odometry error is below $1.1$\,m in both layouts. \textit{Dense Layout} also contains more R2D2 features and yields a lower mean odometry error, consistent with the expected role of visual feature density in supporting odometry quality.

\begin{table}[h]
\centering
\footnotesize
\caption{Results across two real-world layouts.}
\label{tab:real_world_summary}
\setlength{\tabcolsep}{3pt}
\begin{tabular}{lcc}
\toprule
Map & Mean total feature number &  Mean error (m) \\
\midrule
Sparse Layout & 519.7 &  1.0188 \\
Dense  Layout & 969.3 &  0.8204 \\
\bottomrule
\end{tabular}
\end{table}
\begin{figure}[t]
    \centering
    \includegraphics[width=0.49\textwidth]{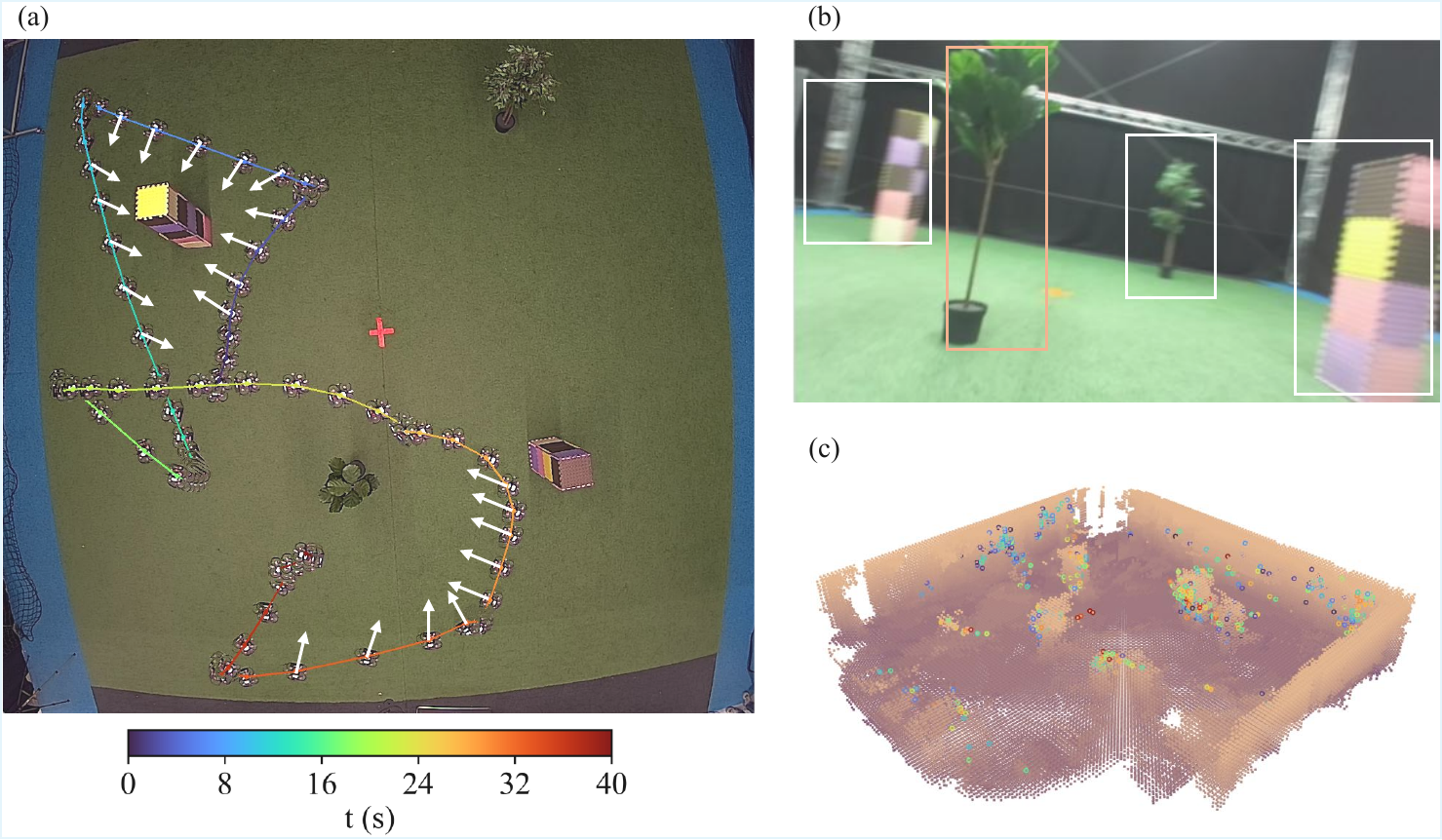}
    \caption{Real-world experiment results in \textit{Dense Layout}. (a) Part of the exploration process from top-down view. White arrows show the camera directions. (b) Camera field of view. Orange denotes visual features currently tracked by the UAV, and white denotes other obstacles. (c) Global feature map overlaid on the occupancy map: feature quality is colour-coded (blue: low score, red: high score), illustrating that high-quality features concentrate on textured obstacles.}
    \label{fig:real_world_experiment}
\end{figure}
Fig.~\ref{fig:real_world_experiment} shows a representative exploration run. In Fig.~\ref{fig:real_world_experiment}a, the UAV prioritises exploration within feature-rich regions, actively tracking features on obstacle surfaces. Fig.~\ref{fig:real_world_experiment}b contrasts currently tracked features (orange rectangles) with untracked scene structure (white). Notably, the tracked features exhibit reduced motion blur, suggesting that yaw trajectory optimisation helps compensate for relative angular motion during flight. Fig.~\ref{fig:real_world_experiment}c depicts the final reconstructed map. 

We further measured the computational time of the main perception-aware modules. On average, the online visibility evaluation, co-visibility sampling, and SQP-based yaw optimisation take $30.262 \pm 16.486$\,ms, $0.0584 \pm 0.0405$\,ms, and $14.039 \pm 14.280$\,ms, showing the perception-aware computations remain lightweight for online planning in current setup.

\section{Conclusions and Future Work}
\label{sec:conclusion}

We presented a hierarchical perception-aware exploration framework that embeds perceptual reasoning directly into frontier selection and yaw planning, without replacing the underlying VIO estimator. Unlike coverage-driven pipelines that treat localisation as a separate, assumed-reliable module, our approach actively steers the UAV toward visually informative subgoals and continuously shapes yaw to maintain stable feature tracks during motion. In simulation across environments with varying texture levels, our method achieves up to $48\%$ higher exploration success rates than coverage-only baselines under strict odometry error thresholds, with the largest gains in the most feature-limited conditions. Real-world trials across two environment layouts validate our approach, with mean odometry errors below $1.1$\,m in predominantly textureless indoor spaces.

The results point to promising opportunities for improvement. Our present framework considers perceptual quality through feature-aware frontier selection and continuous yaw optimisation, while position trajectory generation is still handled separately through safety-driven planning. Future work will thus investigate tighter integration of subgoal selection, position trajectory generation, and yaw angle trajectory, together with richer perceptual objectives that more directly reflect estimator behaviour.
 
%
\bibliographystyle{IEEEtran}
\bibliography{new}

\begin{thebibliography}{10}
\providecommand{\url}[1]{#1}
\csname url@samestyle\endcsname
\providecommand{\newblock}{\relax}
\providecommand{\bibinfo}[2]{#2}
\providecommand{\BIBentrySTDinterwordspacing}{\spaceskip=0pt\relax}
\providecommand{\BIBentryALTinterwordstretchfactor}{4}
\providecommand{\BIBentryALTinterwordspacing}{\spaceskip=\fontdimen2\font plus
\BIBentryALTinterwordstretchfactor\fontdimen3\font minus
  \fontdimen4\font\relax}
\providecommand{\BIBforeignlanguage}[2]{{%
\expandafter\ifx\csname l@#1\endcsname\relax
\typeout{** WARNING: IEEEtran.bst: No hyphenation pattern has been}%
\typeout{** loaded for the language `#1'. Using the pattern for}%
\typeout{** the default language instead.}%
\else
\language=\csname l@#1\endcsname
\fi
#2}}
\providecommand{\BIBdecl}{\relax}
\BIBdecl

\bibitem{yamauchi1998frontier}
B.~Yamauchi, ``Frontier-based exploration using multiple robots,'' in
  \emph{Proc.~of the Intl.~Conf.~on Autonomous Agents}, 1998.

\bibitem{zhou_fuel_2020}
B.~Zhou, Y.~Zhang, X.~Chen, and S.~Shen, ``{FUEL: Fast UAV Exploration Using
  Incremental Frontier Structure and Hierarchical Planning},'' \emph{IEEE
  Robotics and Automation Letters (RA-L)}, vol.~6, no.~2, pp. 779--786, 2021.

\bibitem{zhang_falcon_2024}
Y.~Zhang, X.~Chen, C.~Feng, B.~Zhou, and S.~Shen, ``{FALCON}: Fast autonomous
  aerial exploration using coverage path guidance,'' \emph{IEEE Trans.~on
  Robotics (TRO)}, vol.~41, pp. 1365--1385, 2025.

\bibitem{popovic2024learning}
M.~Popović, J.~Ott, J.~Rückin, and M.~J. Kochenderfer, ``Learning-based
  methods for adaptive informative path planning,'' \emph{Journal on Robotics
  and Autonomous Systems (RAS)}, vol. 179, p. 104727, 2024.

\bibitem{polvara2020next}
R.~Polvara, M.~Fernandez-Carmona, G.~Neumann, and M.~Hanheide,
  ``{Next-best-sense: A multi-criteria robotic exploration strategy for RFID
  tags discovery},'' \emph{IEEE Robotics and Automation Letters (RA-L)},
  vol.~5, no.~3, pp. 4477--4484, 2020.

\bibitem{qin2018vins}
T.~Qin, P.~Li, and S.~Shen, ``{VINS-Mono: A Robust and Versatile Monocular
  Visual-Inertial State Estimator},'' \emph{IEEE Trans.~on Robotics (TRO)},
  vol.~34, no.~4, pp. 1004--1020, 2018.

\bibitem{geneva2020openvins}
P.~Geneva, K.~Eckenhoff, W.~Lee, Y.~Yang, and G.~Huang, ``{OpenVINS: A Research
  Platform for Visual-Inertial Estimation},'' in \emph{Proc.~of the IEEE
  Intl.~Conf.~on Robotics \& Automation (ICRA)}, 2020.

\bibitem{peng2022rwt}
Q.~Peng, X.~Zhao, R.~Dang, and Z.~Xiang, ``{RWT-SLAM: Robust Visual SLAM for
  Weakly Textured Environments},'' in \emph{Proc.~of the IEEE Vehicles
  Symposium (IV)}, 2024.

\bibitem{hardt2020monocular}
A.~Hardt-Stremayr and S.~Weiss, ``Monocular visual-inertial odometry in
  low-textured environments with smooth gradients: A fully dense direct
  filtering approach,'' in \emph{Proc.~of the IEEE Intl.~Conf.~on Robotics \&
  Automation (ICRA)}, 2020.

\bibitem{zhang_perception-aware_2018}
Z.~Zhang and D.~Scaramuzza, ``{Perception-aware Receding Horizon Navigation for
  MAVs},'' in \emph{Proc.~of the IEEE Intl.~Conf.~on Robotics \& Automation
  (ICRA)}.

\bibitem{chen_apace_2024}
X.~Chen, Y.~Zhang, B.~Zhou, and S.~Shen, ``{APACE: Agile and Perception-aware
  Trajectory Generation for Quadrotor Flights},'' in \emph{Proc.~of the IEEE
  Intl.~Conf.~on Robotics \& Automation (ICRA)}, 2024.

\bibitem{yu_perception-aware_2025}
C.~Yu, Z.~Lu, J.~Mei, and B.~Zhou, ``{Perception-aware Planning for Quadrotor
  Flight in Unknown and Feature-limited Environments},'' in \emph{Proc.~of the
  IEEE/RSJ Intl.~Conf.~on Intelligent Robots and Systems (IROS)}, 2025.

\bibitem{cieslewski_rapid_2017}
T.~Cieslewski, E.~Kaufmann, and D.~Scaramuzza, ``Rapid exploration with
  multi-rotors: A frontier selection method for high speed flight,'' in
  \emph{Proc.~of the IEEE/RSJ Intl.~Conf.~on Intelligent Robots and Systems
  (IROS)}.

\bibitem{tao2022seer}
Y.~Tao, Y.~Wu, B.~Li, F.~Cladera, A.~Zhou, D.~Thakur, and V.~Kumar, ``{SEER:
  Safe Efficient Exploration for Aerial Robots using Learning to Predict
  Information Gain},'' in \emph{Proc.~of the IEEE Intl.~Conf.~on Robotics \&
  Automation (ICRA)}, 2023.

\bibitem{yuan2024exploring}
S.~Yuan, H.~U. Unlu, H.~Huang, C.~Wen, A.~Tzes, and Y.~Fang, ``Exploring the
  reliability of foundation model-based frontier selection in zero-shot object
  goal navigation,'' in \emph{Proc.~of the IEEE Conf.~on Computer Vision and
  Pattern Recognition (CVPR)}, 2024.

\bibitem{cao_tare_2021}
C.~Cao, H.~Zhu, H.~Choset, and J.~Zhang, ``{TARE: A Hierarchical Framework for
  Efficiently Exploring Complex 3D Environments},'' in \emph{Proc.~of Robotics:
  Science and Systems (RSS)}, 2021.

\bibitem{dong2025eden}
Q.~Dong, X.~Zhang, S.~Zhang, Z.~Wang, Z.~Ma, and H.~Xi, ``{EDEN: Efficient
  Dual-Layer Exploration Planning for Fast UAV Autonomous Exploration in Large
  3-D Environments},'' \emph{IEEE Trans.~on Industrial Electronics}, pp. 1--11,
  2026.

\bibitem{cao2025dare}
Y.~Cao, J.~Lew, J.~Liang, J.~Cheng, and G.~Sartoretti, ``Dare: Diffusion policy
  for autonomous robot exploration,'' in \emph{Proc.~of the IEEE Intl.~Conf.~on
  Robotics \& Automation (ICRA)}, 2025.

\bibitem{ruckin2022adaptive}
J.~Rückin, L.~Jin, and M.~Popović, ``{Adaptive Informative Path Planning
  Using Deep Reinforcement Learning for UAV-based Active Sensing},'' in
  \emph{Proc.~of the IEEE Intl.~Conf.~on Robotics \& Automation (ICRA)}, 2022.

\bibitem{zhu2024code}
H.~Zhu, T.~Lan, S.~Ma, X.~Zhao, H.~Shang, and R.~Li, ``{CODE: Complete Coverage
  UAV Exploration Planner using Dual-Type Viewpoints for Multi-Layer Complex
  Environments},'' \emph{IEEE Robotics and Automation Letters (RA-L)}, vol.~10,
  pp. 1880--1887, 2024.

\bibitem{geng2025epic}
S.~Geng, Z.~Ning, F.~Zhang, and B.~Zhou, ``{EPIC: A Lightweight LiDAR-Based UAV
  Exploration Framework for Large-Scale Scenarios},'' \emph{IEEE Robotics and
  Automation Letters (RA-L)}, vol.~10, pp. 5090--5097, 2025.

\bibitem{liu2025flare}
X.~Liu, M.~Lin, S.~Li, G.~Xu, Z.~Wang, H.~Wu, and Y.~Liu, ``{FLARE: Fast
  Large-Scale Autonomous Exploration Guided by Unknown Regions},'' \emph{IEEE
  Robotics and Automation Letters (RA-L)}, vol.~10, pp. 12\,197--12\,204, 2025.

\bibitem{costante_perception-aware_2017}
G.~Costante, C.~Forster, J.~Delmerico, P.~Valigi, and D.~Scaramuzza,
  ``Perception-aware path planning,'' \emph{arXiv preprint:1605.04151}, 2017.

\bibitem{bartolomei_perception-aware_2020}
L.~Bartolomei, L.~Teixeira, and M.~Chli, ``{Perception-aware Path Planning for
  UAVs using Semantic Segmentation},'' in \emph{Proc.~of the IEEE/RSJ
  Intl.~Conf.~on Intelligent Robots and Systems (IROS)}, 2020.

\bibitem{bartolomei_semantic-aware_2021}
L.~Bartolomei, L.~Teixeira, and M.~Chli, ``Semantic-aware active perception for
  {UAVs} using deep reinforcement learning,'' in \emph{Proc.~of the IEEE/RSJ
  Intl.~Conf.~on Intelligent Robots and Systems (IROS)}, 2021.

\bibitem{valencia_active_2012}
R.~Valencia, J.~Valls~Miro, G.~Dissanayake, and J.~Andrade-Cetto, ``{Active
  Pose SLAM},'' in \emph{iros}, 2012.

\bibitem{vallve_active_2015}
J.~Vallv{\'e} and J.~Andrade-Cetto, ``{Active Pose SLAM with RRT*},'' in
  \emph{Proc.~of the IEEE Intl.~Conf.~on Robotics \& Automation (ICRA)}, 2015.

\bibitem{pittol_loop-aware_2022}
D.~Pittol, M.~Mantelli, R.~Maffei, M.~Kolberg, and E.~Prestes, ``{Loop-Aware
  Exploration Graph: A Concise Representation of Environments for Exploration
  and Active Loop-Closure},'' \emph{Journal on Robotics and Autonomous Systems
  (RAS)}, vol. 155, p. 104179, 2022.

\bibitem{deshpande_lighthouses_2023}
M.~Deshpande, R.~Kim, D.~Kumar, J.~J. Park, and J.~Zamiska, ``{Lighthouses and
  Global Graph Stabilization: Active SLAM for Low-compute, Narrow-FoV
  Robots},'' in \emph{Proc.~of the IEEE Intl.~Conf.~on Robotics \& Automation
  (ICRA)}, 2023.

\bibitem{maheshwari_region_2025}
M.~Maheshwari, S.~Rabiee, H.~Yin, M.~Labrie, and H.~Liu, ``Region based
  slam-aware exploration: Efficient and robust autonomous mapping strategy that
  can scale,'' \emph{arXiv preprint arXiv:2504.10416}, 2025.

\bibitem{revaud2019neurips}
J.~Revaud, P.~Weinzaepfel, C.~R. de~Souza, and M.~Humenberger, ``{R2D2:}
  repeatable and reliable detector and descriptor,'' in \emph{Proc.~of the
  Conf. on Neural Information Processing Systems (NeurIPS)}, 2019.

\bibitem{mellinger2011minimum}
D.~Mellinger and V.~Kumar, ``Minimum snap trajectory generation and control for
  quadrotors,'' in \emph{Proc.~of the IEEE Intl.~Conf.~on Robotics \&
  Automation (ICRA)}, 2011.

\bibitem{chen2022rast}
G.~Chen, S.~Wu, M.~Shi, W.~Dong, H.~Zhu, and J.~Alonso-Mora, ``Rast: Risk-aware
  spatio-temporal safety corridors for mav navigation in dynamic uncertain
  environments,'' \emph{IEEE Robotics and Automation Letters (RA-L)}, vol.~8,
  no.~2, pp. 808--815, 2022.

\bibitem{liu2017planning}
S.~Liu, M.~Watterson, K.~Mohta, K.~Sun, S.~Bhattacharya, C.~J. Taylor, and
  V.~Kumar, ``Planning dynamically feasible trajectories for quadrotors using
  safe flight corridors in 3-d complex environments,'' \emph{IEEE Robotics and
  Automation Letters (RA-L)}, vol.~2, no.~3, pp. 1688--1695, 2017.

\bibitem{johnson2007nlopt}
S.~G. Johnson, ``The nlopt nonlinear-optimization package,'' 2007.

\bibitem{kraft1994tomp}
D.~Kraft, ``Algorithm 733: Tomp--fortran modules for optimal control
  calculations,'' \emph{{ACM} Transactions on Mathematical Software}, vol.~20,
  pp. 262--281, 1994.

\bibitem{foehn_agilicious_2022}
P.~Foehn, E.~Kaufmann, A.~Romero, R.~Penicka, S.~Sun, L.~Bauersfeld,
  T.~Laengle, G.~Cioffi, Y.~Song, A.~Loquercio, and D.~Scaramuzza,
  ``Agilicious: Open-source and open-hardware agile quadrotor for vision-based
  flight,'' \emph{Science Robotics}, vol.~7, no.~67, 2022.

\end{thebibliography}

\newpage

\vfill

\end{document}